\documentclass{article}

\usepackage{PRIMEarxiv}

\usepackage[utf8]{inputenc} % allow utf-8 input
\usepackage[T1]{fontenc}    % use 8-bit T1 fonts
\usepackage{hyperref}       % hyperlinks
\usepackage{url}            % simple URL typesetting
\usepackage{booktabs}       % professional-quality tables
\usepackage{amsfonts}       % blackboard math symbols
\usepackage{nicefrac}       % compact symbols for 1/2, etc.
\usepackage{microtype}      % microtypography
\usepackage{lipsum}
\usepackage{fancyhdr}       % header
\usepackage{graphicx}       % graphics
\graphicspath{{media/}}     % organize your images and other figures under media/ folder

\usepackage{amsmath}
\usepackage{amssymb}
\usepackage{color}
\usepackage{bbding}

\title{Low-Light Image Enhancement\\ in the Frequency Domain}

\author{Hao Chen\\
Sun Yat-sen University\\
{\tt\small chenh366@mail2.sysu.edu.cn }
\and Zhi Jin\\
Sun Yat-sen University\\
{\tt\small jinzh26@mail.sysu.edu.cn}}

\begin{document}
\maketitle

% ABSTRACT
\begin{abstract}

Decreased visibility, intensive noise, and biased color are the common problems existing in low-light images. These visual disturbances further reduce the performance of high-level vision tasks, such as object detection, and tracking. To address this issue, some image enhancement methods have been proposed to increase the image contrast. However, most of them are implemented only in the spatial domain, which can be severely influenced by noise signals while enhancing. Hence, in this work, we propose a novel residual recurrent multi-wavelet convolutional neural network ({\bf R2-MWCNN}) learned in the frequency domain that can simultaneously increase the image contrast and reduce noise signals well. This end-to-end trainable network utilizes a multi-level discrete wavelet transform to divide input feature maps into distinct frequencies, resulting in a better denoise impact. A channel-wise loss function is proposed to correct the color distortion for more realistic results. Extensive experiments demonstrate that our proposed R2-MWCNN outperforms the state-of-the-art methods quantitively and qualitatively.  
%The code can be found here: \href{https://github.com/chqwer2/How_to_see_in_the_Dark}{\textcolor{blue}{https://github.com/chqwer2/How\_to\_see\_in\_the\_Dark}}

\end{abstract}

% insufficient illumination
% captured under adverse illumination conditions
%-------------------------------------------------------------------------
% Introduction
\section{Introduction}
\label{sec:intro}

\begin{figure}
% \hspace{-0.2cm}
\centering
\begin{tabular}{ccc}
\hspace{-0.25cm}
\includegraphics[width=5.2cm]{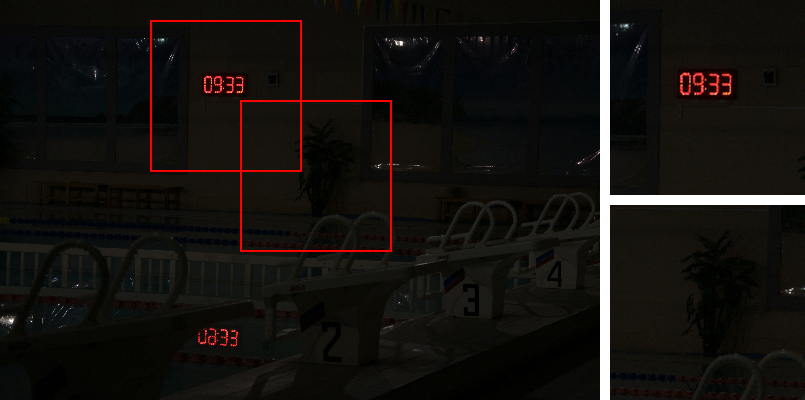}&
{\includegraphics[width=5.2cm]{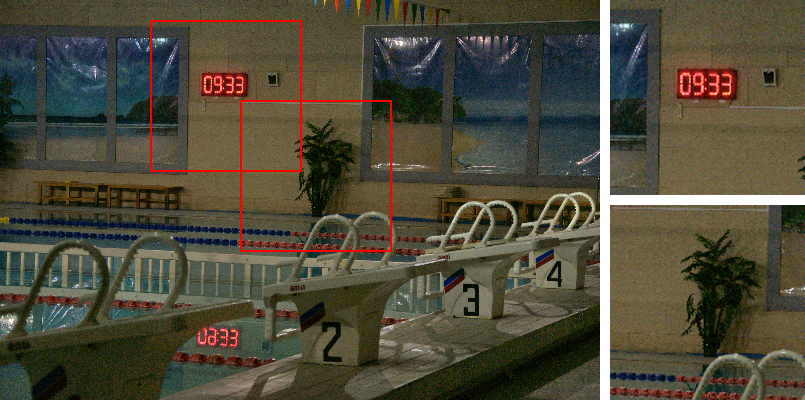}}&
{\includegraphics[width=5.2cm]{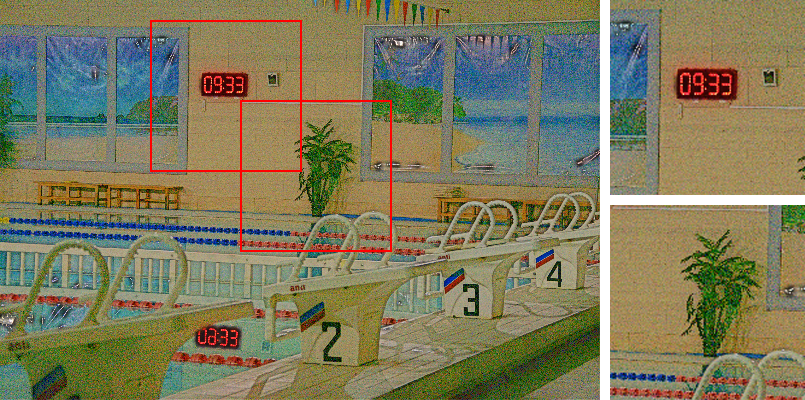}}\\

\hspace{-0.25cm}
(a) Original&(b) LIME&(c) RetinexNet
\end{tabular}
\hfill
\begin{tabular}{ccc}
\hspace{-0.25cm}
{\includegraphics[width=5.2cm]{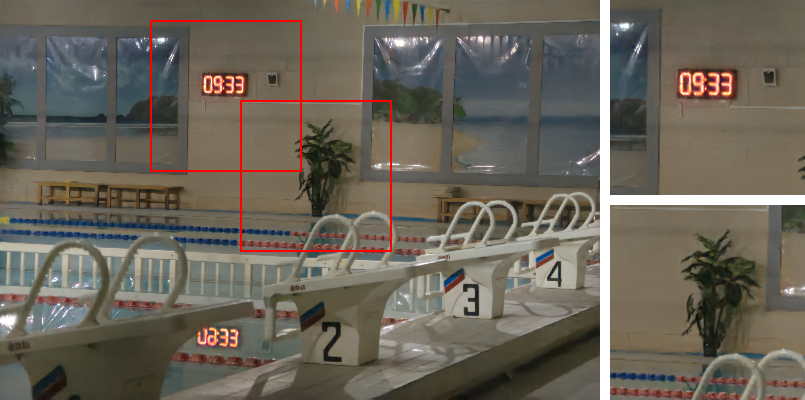}}&
{\includegraphics[width=5.2cm]{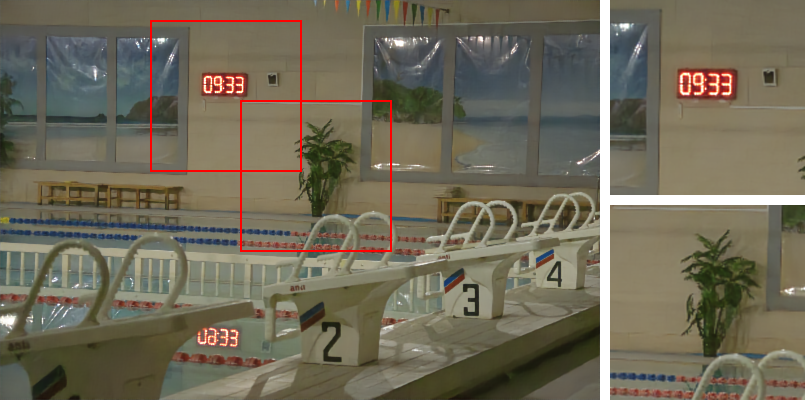}}&
{\includegraphics[width=5.2cm]{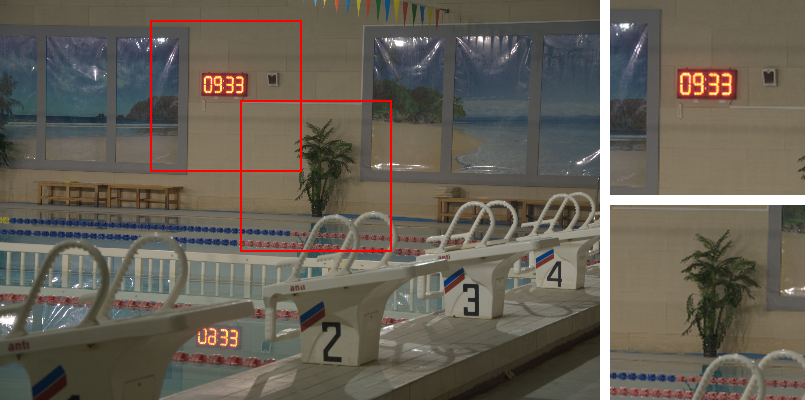}}\\

\hspace{-0.25cm}
(d) DRBN&(e) R2-MWCNN&(f) Reference
\end{tabular}
\caption{The enhanced results in {\em LOL-Real} dataset \cite{yang2021sparse} of different methods (PSNR/SSIM). (a) Low-light input. (b) LIME \cite{LIME} (19.2096/0.5778). (c) RetinexNet \cite{LOL} (17.0594/0.4011). (d) DRBN \cite{yang2020fidelity} (25.8050/0.9273). (e) R2-MWCNN (27.8149/0.9378).  (f) Normal light reference. The proposed R2-MWCNN well restores high quality images from real low-light inputs in the {\em LOL-Real} dataset.}
\label{fig1}
\end{figure}

Low-light image enhancement is critical for many vision tasks, including recognition \cite{wade1998fast} and nighttime autonomous driving \cite{rashed2019fusemodnet}. However, low-light scenarios introduce a succession of visual degradations, e.g. diminished visibility, intensive noise and color distortion in captured images. 
These images suffer from information loss and are likely to undermine the performance of algorithms that are primarily designed for high-visibility conditions. Therefore, to make the buried information visible, as shown in Figure \ref{fig1}, low-light image enhancement is expected. However, it is non-trivial to enhance low-light images, as the noise and color distortion are easily be amplified unconsciously.

Many researchers have focused on addressing these issues over the last few decades, and many effective algorithms have been proposed. Traditional low-light image enhancement algorithms can be broadly classed as Histogram Equalization-related \cite{HE} and Retinex theory-related \cite{Land} methods. 
Histogram Equalization (HE) and its derivatives \cite{cheng2004simple, abdullah2007dynamic}, which mainly focus on enhancing the constrast of images at the pixel level, can unintendedly amplify blind noise while restoring illumination. The Retinex theory-based algorithms \cite{wang2013naturalness, LOL} decompose image into a constant reflectance layer and a variant illumination layer, with the purpose of restoring the illumination layer of the image independently. In practice, however, reflectance and illumination layers are roughly estimated, which can cause visible color distortion and further decrease visual quality.

%  It also focuses on suppressing noise and preserving structure details within the images.

The recent blossoming of deep learning has inspired the development of low-light enhancement. Some approaches \cite{cai2018learning, LOL} are built on the Retinex theory, while some others design CNNs \cite{chen2018learning, guo2020zero} and GANs \cite{liu2021pd} to enhance low-light images directly. However, the shared problem is that these methods are mainly designed in the pixel domain, which suits human perception rather than machine vision. 

%Thus, the good visual quality cannot naturally achieved. 
%the drawback of pixel domain-based methods is that 
Precisely, for computer vision, the random noise carried by low-light images is hardly to be separated in the pixel domain. Most of the existing denoising algorithms perform an individual filtering process or noise estimation elimination, and can efficiently reduce the noise levels. However, the images are either blurred or over-smoothed, resulting in unsatisfactory details and visual quality \cite{agaian2001transform, bijalwan2012wavelet}. Therefore, although recalling the visibility in the pixel domain is the most intuitive and rather effective way, their capabilities for noise suppression and detail restoration are limited.

% there is still a lot room to improve. For
% spectral information
%seeing this tremendous advancement in low-light improvement techniques
% {\em  
Considering the separability between noise and structural information in the frequency domain \cite{bijalwan2012wavelet}, in this paper, we propose a frequency-based architecture to address the problems of enhancing low-light images in the pixel domain. The core of our architecture is a novel residual recurrent multi-wavelet convolutional neural network (R2-MWCNN).
The Discrete Wavelet Transform (DWT) and Inverse Discrete Wavelet Transform (IDWT) \cite{liu2018} are introduced to replace commonly used pooling and up-sampling operations and exploit multi-scale frequency-based features for denoising and detail restoration. We also propose the Multi-level Short-cut Connections (MSC) module to select and transmit low-level semantic information adaptively for detail recovery.  
Moreover, since enhanced low-light images usually suffer from severe color distortion, a channel-wise loss function is proposed to aid the retrieval of pleasing visual quality. Overall, our contributions are as follows:
\begin{itemize}
	\item {We propose a novel end-to-end frequency-based architecture for enhancing low-light images. It integrates illumination enhancement, color restoration and denoising, and can improve both the objective and subjective image quality.}
	\item {We propose an effective low-light image enhancement network with the proposed MSC modules to adaptively select the high-/low-level semantic features for satisfactory detail restoration. DWT and IDWT are introduced to exploit multi-scale frequency-based features, which serves as an appropriate basis for separating noise from the structural information.}
	\item {A novel channel-wise loss function is designed to handle the pixel-intensity distribution of each color channel in order to overcome color distortion and improve the visual quality. Extensive experiments show that our framework outperforms the existing state-of-the-art methods both quantitatively and qualitatively.}
\end{itemize}

\section{Related Work}
Low-light enhancement techniques can be classified into classic or deep learning-based algorithms according to whether a data-driven deep network is adopted.

{\bf Classic enhancement algorithms.  }
Histogram equalization and its derivatives are one of the classic algorithm categories. CLAHE \cite{clahe} attempts to alleviate noise amplification by limiting the range of contrast. BPDHE \cite{ibrahim2007brightness} aims to maintain brightness while dynamically correcting contrast. LDR \cite{lee2013contrast} proposes investigating the layered difference representation of 2-D histograms in order to enhance contrast by emphasizing gray-level differences between adjacent pixels. DHECE \cite{nakai2013color} combines differential gray-levels histogram equalization of saturation and intensity to generate appropriate high-contrast results.

Another traditional low-light enhancement category is built based on the Retinex theory, which assumes an image consists of a constant reflectance layer and a variant illumination layer. Hence, if these two layers can be successfully separated and enhanced properly, the low-light image can be enhanced well by combining them.
SSR \cite{jobson1997properties} and MSR \cite{rahman1996multi} are typical Retinex theory-based enhancement algorithms that attempt to rectify the illumination map for brightness enhancement. AMSR \cite{lee2013adaptive} proposes a weighting strategy for each MSR output. MSRCR \cite{jobson1997multiscale} tries to enhance reflectance estimation with a color restoration factor. RRM \cite{li2018structure} designs an optimization function to constrain noise in illumination estimates. SRIE \cite{fu2016weighted} employs a weighted vibrational model to estimate reflectance and illumination. LIME \cite{LIME} refines the maximum-approximated illumination and recomposes the enhancement result in accordance with Retinex theory.

However, these methods heavily rely on manually designed parameters that suit different low-light images. Hence, they are not flexible and easily amplify noise and color distortion, causing decreased visual quality.

{\bf Deep learning-based methods. }  Recently, deep learning has stimulated tremendous progress in low-light image enhancement.
For example, LLNet \cite{llnet} is the first deep learning trial in low-light image enhancement, proposing an auto-encoder for contrast enhancement and denoising. MSR-net \cite{shen2017msr} is the earliest implementation of the Retinex theory-based supervised CNN. EnlightenGAN \cite{enlightengan} is a successful trial of Generative Adversarial Nets (GANs) that can adaptively learn from unpaired low/normal-light images. Retinex-Net \cite{LOL} is designed to integrate deep learning with Retinex theory, which can decompose the image to enhance the illumination and reflectance, respectively. DeepUPE \cite{wang2019} presents a neural network that enhances under-exposure images by learning complicated photography adjustments. SICE \cite{SICE} includes an end-to-end CNN to adapt the contrast of multi-exposure input images. MBLLEN \cite{MBLLEN} adopts three modules within a single network, improving the output of each extraction layer and combining them together to create the final enhanced image. DRBN \cite{yang2020fidelity} proposes a semi-supervised learning framework to enhance low-light images from coarse to fine.

Although these methods have demonstrated outstanding low-light enhancement performance, due to the limitations of the pixel domain, blind noise can not be effectively eliminated from the images. Blurring and color distortion, which are caused by spatial denoising, continue to obstruct practical application. To overcome this shared limitation, R2-MWCNN employs an invertible wavelet transform \cite{pan1999two} to perform denoising in the frequency domain so that to effectively recover sharp structures from degraded observation. In real-world scenes, the image information has a certain structure, but the noise is random without a specific structural property. Hence, the introduction of the wavelet transform allows the network to separate the image structural information and the noise, which aids in preserving pleasing details during denoising. Moreover, a novel channel-wise loss is proposed to correct the common color distortion problem in low-light enhancement.

\begin{figure}[t]
\includegraphics[width=17.2cm]{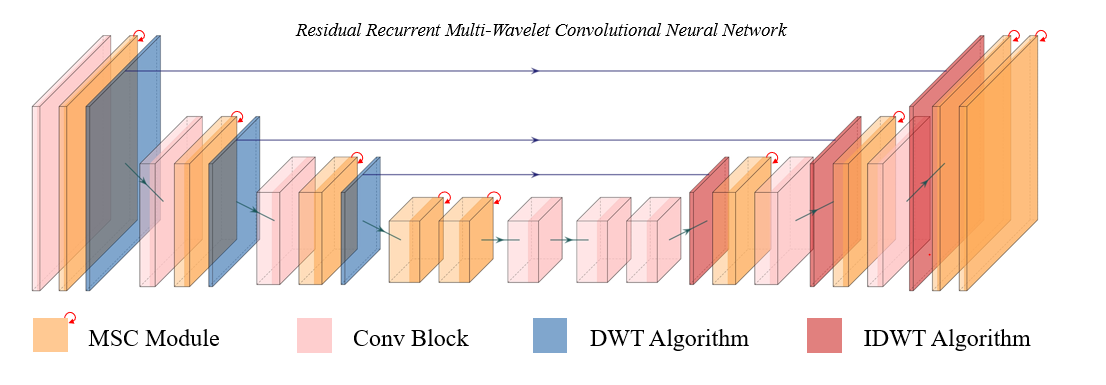}
\caption{The end-to-end architecture of the Residual Recurrent Multi-Wavelet Convolutional Neural Network (R2-MWCNN) with U-Net  \cite{ronneberger2015u}, wavelet transform and Multi-level Short-cut Connections module (red swirling arrow).}
\label{Structure}
\end{figure}

\section{Methodology}
\subsection{R2-MWCNN Network}
Our network comes from two inspirations. First, it is easier to remove noise from image structural information in the frequency domain and to estimate the global illumination from the low-frequency features \cite{xu2020learning}. Second, it is known that as the network depth increases, the model can extract higher-level semantic information from the image. Nevertheless, with the common employment of down-sampling operations, the original low-level structural information may be lost as the network deepens. Hence, as shown in the Figure \ref{Structure}, the proposed R2-MWCNN is built with the U-Net structure \cite{ronneberger2015u}, wavelet transform and several MSC modules. The MSC module is employed by replacing normal covolutional blocks, which can transmit low-level graphic information into deeper network layers without requiring additional computational cost. Our R2-MWCNN comprises a contracting subnetwork and an expanding subnetwork. In the contracting subnetwork, the invertible DWT is introduced to replace the frequently used pooling operation and perform space-to-frequency tranform. In the expansion subnetwork, IDWT is presented to restore all the information that is kept by DWT to high resolution in the pixel domain.

\begin{figure}
\centering
\begin{tabular}{cc}
{\includegraphics[width=8cm]{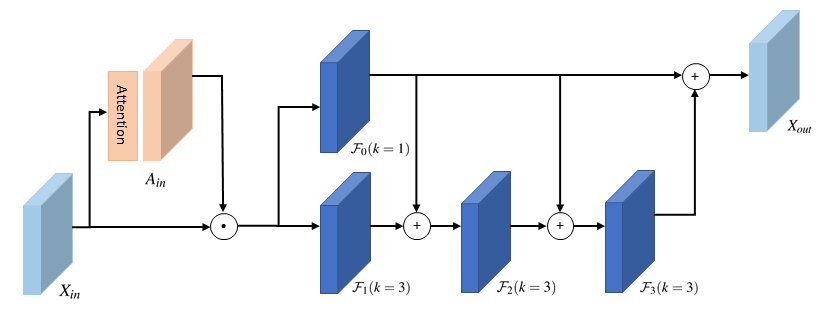}}&
{\includegraphics[width=8cm]{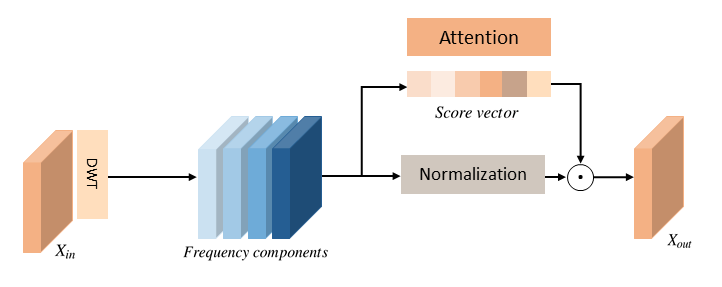}}\\
(a) MSC Module with Attention Map&(b) DWT with Attention Mechanism 
\end{tabular}

\caption{The MSC module with attention map utilizes several shortcut connections and multiple add operations to maintain low-level original graphic information. The DWT with attention mechanism performs space-to-frequency transform, which enables the model to separate noisy signals from structural information.}
\label{block}
\end{figure}

{\bf MSC module.} Inspired by recurrent convolutional layer \cite{alom2018recurrent}, we propose the MSC module, as shown in Figure \ref{block} (a). The goal of the MSC module is to preserve and transmit low-level features into subsequent layers without requiring additional processing resources, so as to restore satisfactory details. Given the input features $X_{in} \in \mathcal{R}^{H \times W \times C}$ and convolutional operation $\mathcal{F( \cdot )}$, the MSC module first calculates the pixel-wise contrast-aware attention map $C_a$ \cite{lu2012learning} to select features by:
\begin{equation}
\begin{split}
C_{a} &= sigmoid(\mathcal{F}_{map}(X_{in})) ,\\
X_a & = C_{a} \cdot A_{in} .
\end{split}
\end{equation}

Next, considering the information loss as the network deepens, several shortcut connections are conducted between the input layer and the subsequent convolutional layers in the MSC module to adaptively select low-level graphic information. The shortcut connection \cite{venables2013modern} performs an add operation between the identity mapping and the output of previously stacked features, which can be calculated by $\mathcal{F}(x_1)+x_2$. Considering that the output of the $l^{th}$ layer in the MSC module is $X_{l+1}$, which can be calculated accordingly:
\begin{equation}
\begin{split}
X_{l+1} &= \mathcal{F}_l (X_l + \mathcal{F}_0(X_a)) ,
\end{split}
\end{equation}
where $\mathcal{F}_0( \cdot )$ represents $1 \times 1$ convolution of input $X_a$ to match the output dimensions.

{\bf Wavelet Transform.} To guarantee that all the information can be well-preserved, in the R2-MWCNN, we utilize DWT and IDWT to replace the commonly-used downsampling and upsampling. The DWT can be viewed as a top-down collection of several filters that can theoretically extract both frequency and location features, shown in Figure \ref{block} (b), while IWT is employed to substitute the transpose convolutional layer to integrate low resolution features maps into recovered high resolution ones. 

In practice, given the trade-off between fidelity gain and computational cost, a two-layer DWT with four sub-band outputs ($X_{LL}, X_{LH}, X_{HL}$ and $X_{HH}$) is adopted \cite{liu2018}, where the subscript index is an ordered representation of filter processing. For example, assuming the input sample is $X_{in} \in \mathcal{R}^{H \times W \times C}$,  $X_{LL}$ is the output of $X$ that passes through two low-pass filters, which can be calculated as below:
\begin{equation}
 X_{LL} =X_{in}  \otimes f_{LL} ,
\end{equation}
where $f_{LL}$ can be defined as:
\begin{equation}
 f_{LL} = \begin{bmatrix} 1 & 1 \\ 1 & 1 \end{bmatrix} .
\end{equation}

Analogously, the rest of the filter definitions are shown as follows:
\begin{equation}
f_{LH} = \begin{bmatrix} -1 & -1 \\ 1 & 1 \end{bmatrix},\quad  f_{HL} = \begin{bmatrix} -1 & 1 \\ -1 & 1 \end{bmatrix},\quad f_{HH} = \begin{bmatrix} 1 & -1 \\ 1 & -1 \end{bmatrix}\quad .
\end{equation}

As can be seen, the pixel-intensity of the four sub-band outputs is substantially distinctive. Hence, we normalize the value range to $[0, 1]$ to ease computational burden. Finally, the attention mechanism \cite{fukui2019attention} is introduced to adaptively scale the feature maps from different sub-bands in a channel-wise manner.

% both subjective and objective 
\subsection{Loss Function}
The Mean Square Error (MSE), also known as L2 loss, is a commonly used error measurement for low-light image enhancement. Nevertheless, L2 loss has been proven to be insufficient for recovering favourable details and can even induce blur at the object edges \cite{sara2019image}.  In order to finely recover the visual details of the image, we use multiple losses to measure differences in various aspects of the enhanced image and the corresponding ground truth. The whole loss function for training consists of four parts:
\begin{equation}
Loss = L_{pixel} + L_{global} + L_{edge} + L_{channel}
\end{equation}

{\bf Pixel-wise Loss.} The pixel-wise loss, indicated as $L_{pixel}$, is a variant of the Smooth L1 loss. The wider gap between blind noise and structural information in Smooth L1 loss compared to L2 loss, makes the network more focused on denoising. Experiments reveal that our network performs well in dark regions yet over-enhances bright regions. After conducting several trials, the images are divided into bright and dark region with a 3:7 area ratio. Different weights $w_{1}$ and $w_{2}$ are assigned to provide a better approximation of the bright region.
\begin{equation}
\begin{split}
 L_{pixel} &=  w_{1} \cdot  \frac{1 }{n_L m_L} \sum_{i=1}^{n_L} \sum_{j=1}^{m_L} Smooth_{L1} (E_L(i,j), G_L(i, j)) \\
   & + w_{2} \cdot \frac{1}{n_H m_H}  \sum_{i=1}^{n_H} \sum_{j=1}^{m_H} Smooth_{l1} (E_H(i,j), G_H(i, j)) , 
\end{split}
\end{equation}
where $E_H$ and $G_H$ represent the bright region of the enhanced image and the ground truth reference, and $E_L$ and $G_L$ are the rest of the image.

{\bf Global-wise Loss.} The global-wise loss is a combination of structure loss and perceptual loss \cite{johnson2016perceptual}. Structural loss is calculated using Structure Similarity (SSIM) \cite{wang2004image}, which is designed to assess the global variations in contrast, brightness, and structure between the enhanced image and the normal image from a global perspective. The perceptual loss is offered to primarily concerned with high-level context via the employment of feature-extraction VGG network \cite{Simonyan2015VeryDC}. If the enhanced results and the original images are close, the VGG extractor outputs should be similar as well. Our global-wise loss is calculated as follows:
\begin{equation}
\begin{split}
 L_{global} = w_{3} \cdot \frac{1}{C_{i,j} H_{i,j} W_{i,j}} ||\phi_{i,j}(E) - \phi_{i,j}(G)||_2^2
 -  w_{4} \cdot \frac{(2 \mu_E \mu_G + C_1)(2\sigma_{EG} + C_2)}{(\mu_E^2 + \mu_G^2 + C_1)   (\sigma_{E}^2 + \sigma_G^2 + C_2) } ,
\end{split}
\end{equation}
where $E$ and $G$ represent enhanced image and ground truth, respectively, and $\phi_{i,j}(\cdot)$  refers to the feature extractor of the $i_{th}$ block $j_{th}$ layer in the VGG network. $C_{i,j}$, $W_{i,j}$ and $H_{i,j}$ describe the 3-dimensions of feature maps extracted from the VGG-19 network. Towards the structure loss, $\mu_x$ and $\sigma_x$ are average and variance, respectively, while $\sigma_{xy}$ represents covariance between images. $C$ is a small constant to prevent division by zero.

{\bf Edge Loss.} Normally, denoising operations can introduce severe blurring. Words and object edges could be blurred after enhancement, which reduces the practicality of the low-light enhancement algorithms. With respect to that visual deterioration, the edge loss is utilized to evaluate the generated edges. In practice, the Sobel gradient is utilized as the edge information in the images. Two Sobel calculators, $S_x$ and $S_y$, with x-axis and y-axis orientations, respectively, are used to estimate the gradient between two adjacent pixels. Thus, the edge loss between $E$ and $G$ can be calculated as follows:
%\begin{equation}
%\begin{split}
%S_x = \begin{bmatrix} -1 & 0 & 1 \\ -2 & 0 & 2 \\ -1 & 0 & 1 \end{bmatrix},\quad  S_y = \begin{bmatrix} -1 & -2 & -1 \\ 0 & 0 & 0 \\ 1 & 2 & 1 \end{bmatrix} ,
%\end{split}
%\end{equation}
 \begin{equation}
\begin{split}
L_{edge} = ||S_x \otimes E - S_x \otimes G||_1 + ||S_y \otimes E - S_y \otimes G||_1 .
\end{split}
\end{equation}

{\bf Channel-wise Loss.} The previous methods generally use element-wise differences like L2 loss to measure the color fidelity. However, these methods often have color channel prejudice, resulting in generating color-biased images. The proposed channel-wise loss can make the network learn the pleasing global color distribution from bright images, which can guide low-light image enhancement. Therefore, satisfactory colors can be achieved. Firstly, Gaussian blur is used to eliminate image textures with kernel $K_G$:
 \begin{equation}
\begin{split}
K_G(k, l) =  A \cdot exp(- \frac{(k-\mu_x)^2}{2\sigma_x} -   \frac{(l-\mu_y)^2}{2\sigma_y}  ) ,
\end{split}
\end{equation}
where A is defined as 0.2, and Gaussian blurred image of $E$ is referred to as $E_b$:
\begin{equation}
\begin{split}
E_b(i, j) = \sum_i^{W-k} \sum_j^{H-l} E(i+k, j+l) \cdot K_G(k, l) ,
\end{split}
\end{equation}
the operation towards $G$ is the same. Hence, the channel-wise loss can be calculated by:
\begin{equation}
\begin{split}
L_{channel} = \sum_h^{C} ||\sum_i^{W} \sum_j^{H} E_b(i, j, h) - \sum_i^{W} \sum_j^{H} G_b(i, j, h))||_1 .
\end{split}
\end{equation}
The channel-wise loss can suppress biased color, which is typically produced by over or under enhancement of one or more of the RGB channels. Therefore, the network that is guided by this channel-wise loss can retrieve more real colors.

% Here
\section{Experiment}
\subsection{Implementation Setting}
We have implemented the proposed model in the Tensorflow framework \cite{abadi2016tensorflow}. The R2-MWCNN is optimized by the ADAM optimizer for 300 epoches with randomly initialized parameters. The initial learning rate and batch size for training are set to 2$e^{-4}$ and 2. In order to reduce the computational cost, the input image values are scaled to $[0, 1]$. The learning rate is automatically decayed by the Tensorflow ReduceLROnPlateau with a patience of 10 and a decrease factor of 0.2. Rotation and flip based data augmentation have been adopted. The entire training process takes around 1 day to converge with the NVIDIA Tesla P40.

% here
% R2-MWCNN is a robust framework that can make use of  both frequency and location characteristics in images. 
\begin{figure}
\centering
\begin{tabular}{cccccc}
\hspace{-0.3cm}
{\includegraphics[width=2.68cm]{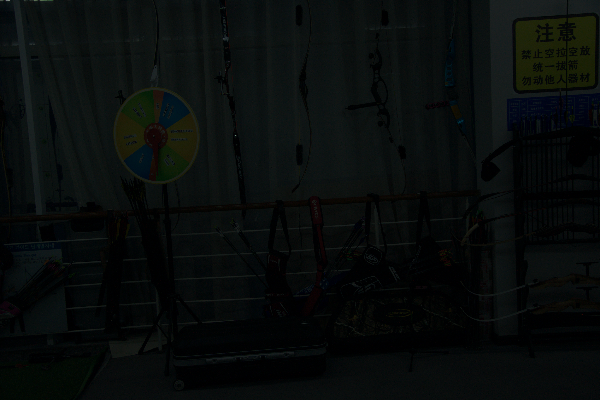}}&
\hspace{-0.42cm}
{\includegraphics[width=2.68cm]{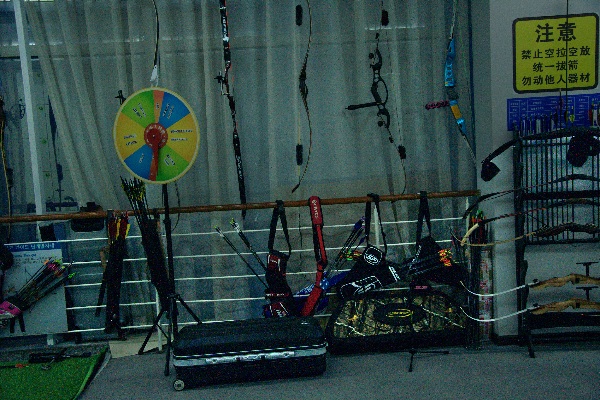}}&
\hspace{-0.42cm}
{\includegraphics[width=2.68cm]{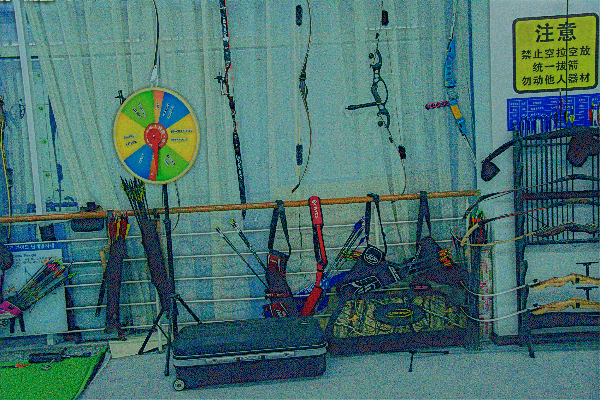}}&
\hspace{-0.42cm}
{\includegraphics[width=2.68cm]{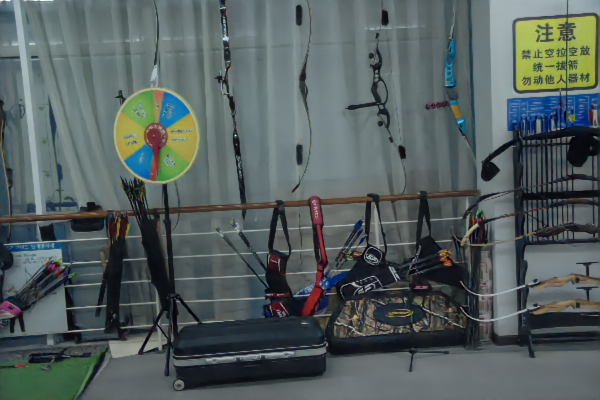}}&
\hspace{-0.42cm}
{\includegraphics[width=2.68cm]{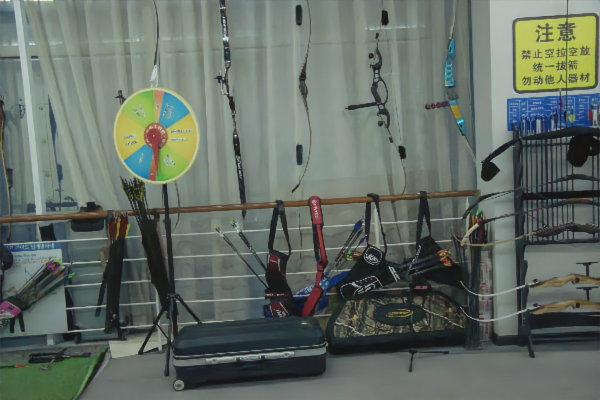}}&
\hspace{-0.42cm}
{\includegraphics[width=2.68cm]{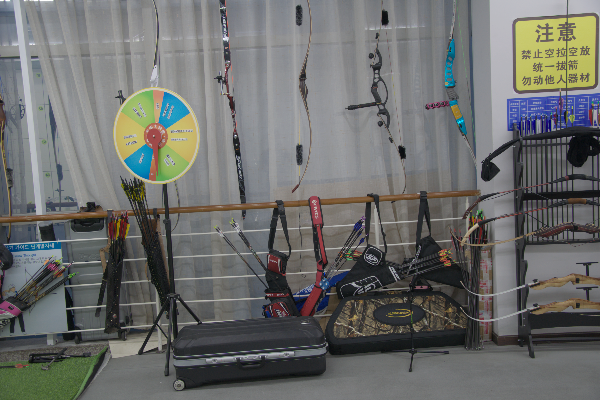}}\\
\end{tabular}
\hfill
\begin{tabular}{cccccc}
\hspace{-0.3cm}
{\includegraphics[width=2.68cm]{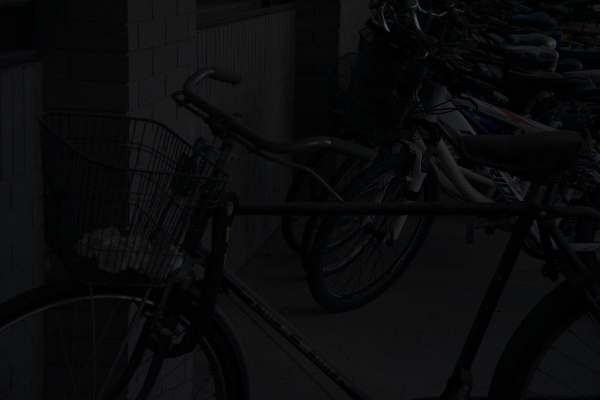}}&
\hspace{-0.42cm}
{\includegraphics[width=2.68cm]{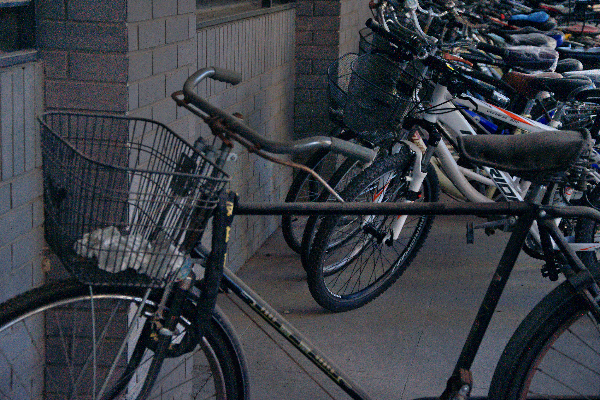}}&
\hspace{-0.42cm}
{\includegraphics[width=2.68cm]{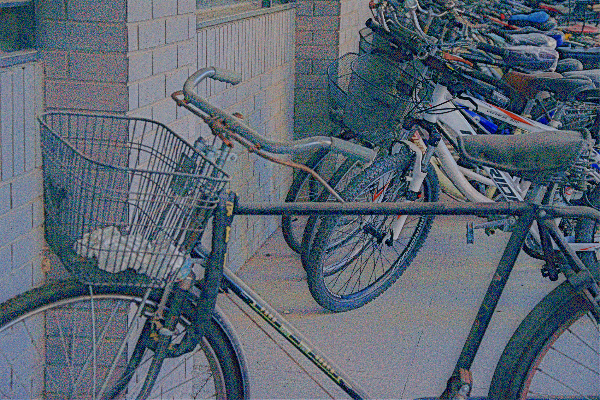}}&
\hspace{-0.42cm}
{\includegraphics[width=2.68cm]{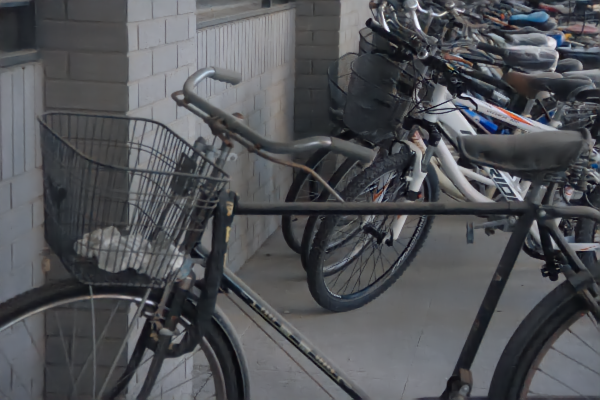}}&
\hspace{-0.42cm}
{\includegraphics[width=2.68cm]{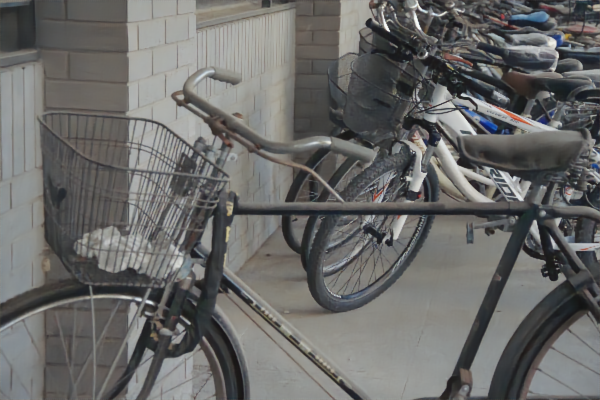}}&
\hspace{-0.42cm}
{\includegraphics[width=2.68cm]{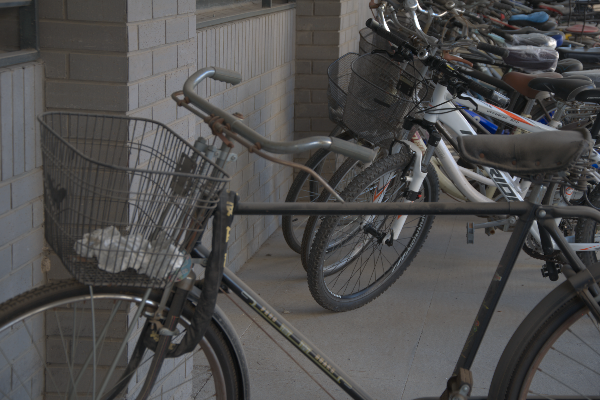}}\\
\end{tabular}
\hfill
\begin{tabular}{cccccc}
\hspace{-0.3cm}
{\includegraphics[width=2.68cm]{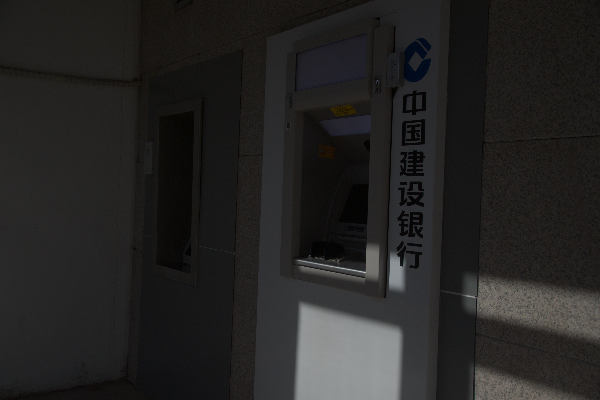}}&
\hspace{-0.42cm}
{\includegraphics[width=2.68cm]{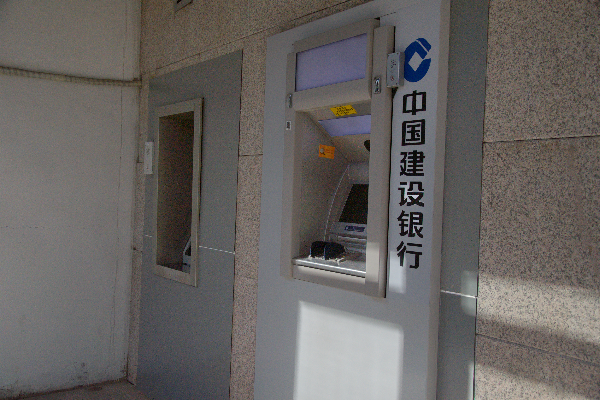}}&
\hspace{-0.42cm}
{\includegraphics[width=2.68cm]{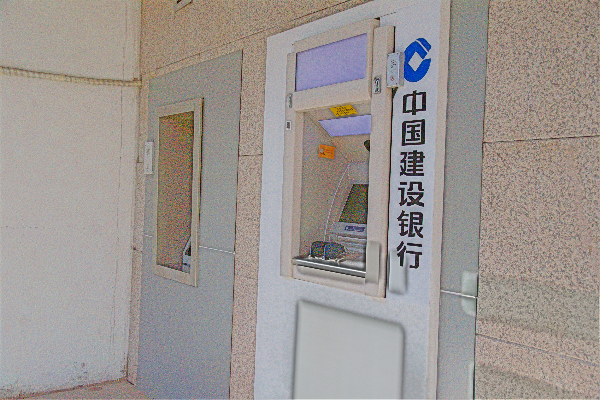}}&
\hspace{-0.42cm}
{\includegraphics[width=2.68cm]{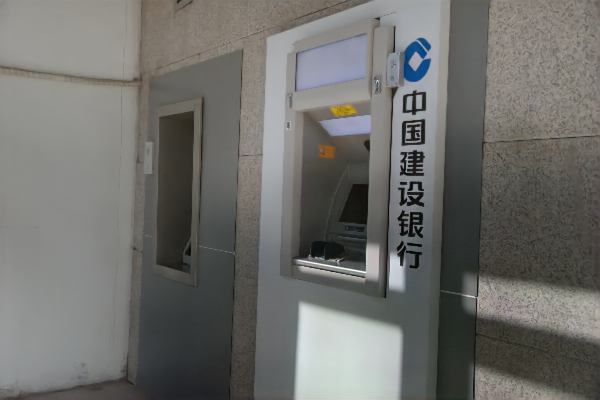}}&
\hspace{-0.42cm}
{\includegraphics[width=2.68cm]{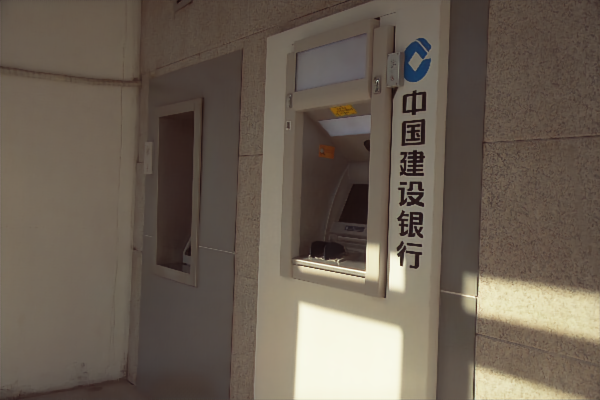}}&
\hspace{-0.42cm}
{\includegraphics[width=2.68cm]{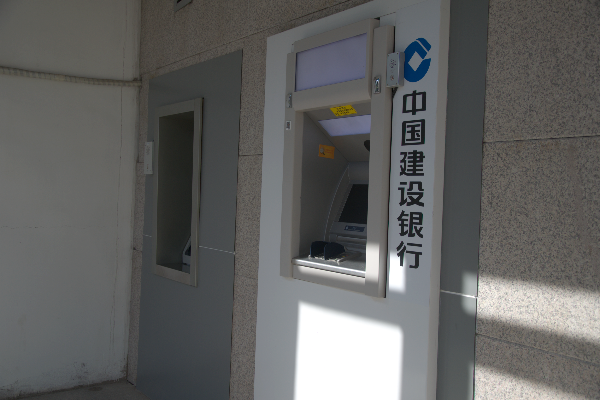}}\\
\end{tabular}
\hfill  %764
\begin{tabular}{cccccc}
\hspace{-0.3cm}
{\includegraphics[width=2.68cm]{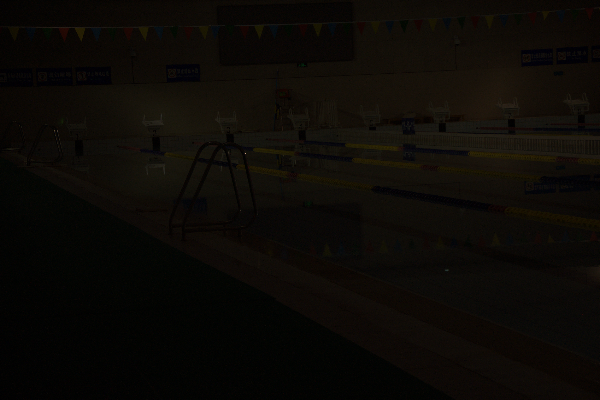}}&
\hspace{-0.42cm}
{\includegraphics[width=2.68cm]{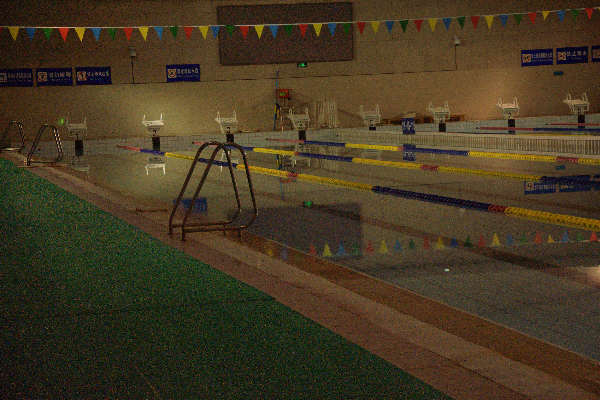}}&
\hspace{-0.42cm}
{\includegraphics[width=2.68cm]{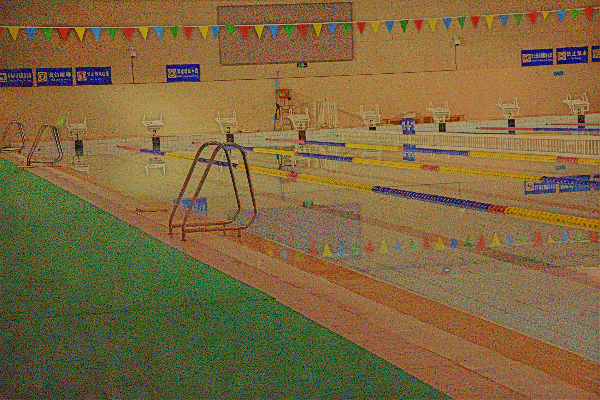}}&
\hspace{-0.42cm}
{\includegraphics[width=2.68cm]{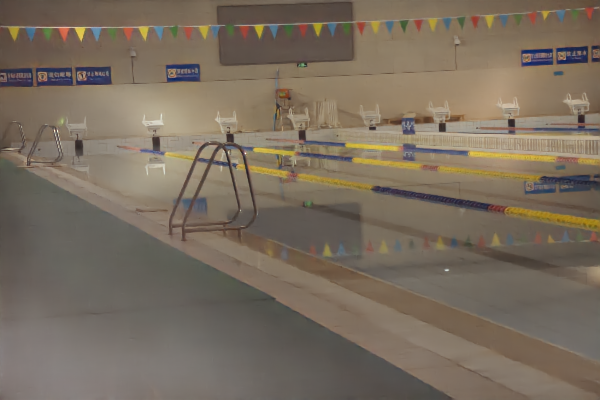}}&
\hspace{-0.42cm}
{\includegraphics[width=2.68cm]{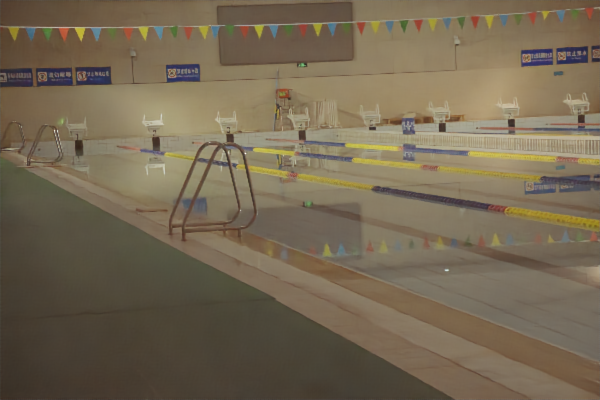}}&
\hspace{-0.42cm}
{\includegraphics[width=2.68cm]{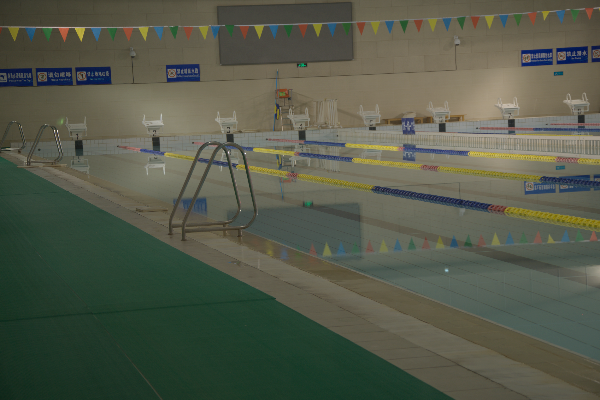}}\\
\end{tabular}
\hfill
\begin{tabular}{cccccc}
\hspace{-0.3cm}
{\includegraphics[width=2.68cm]{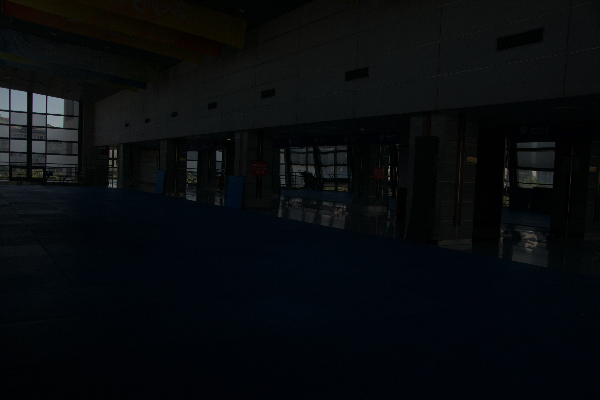}}&
\hspace{-0.42cm}
{\includegraphics[width=2.68cm]{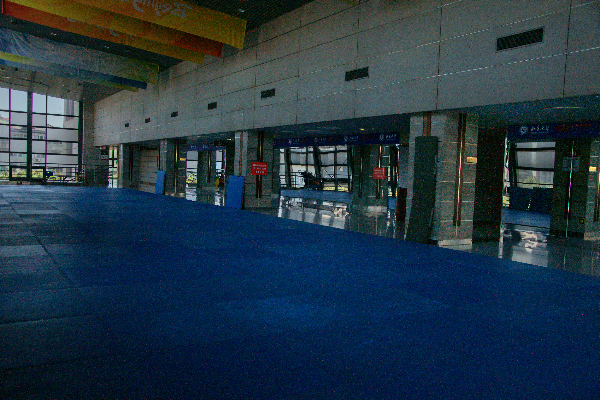}}&
\hspace{-0.42cm}
{\includegraphics[width=2.68cm]{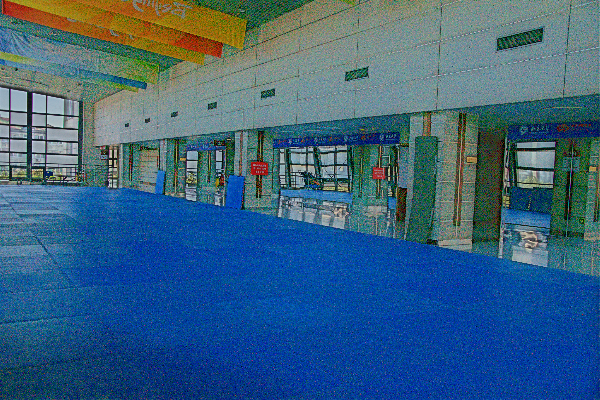}}&
\hspace{-0.42cm}
{\includegraphics[width=2.68cm]{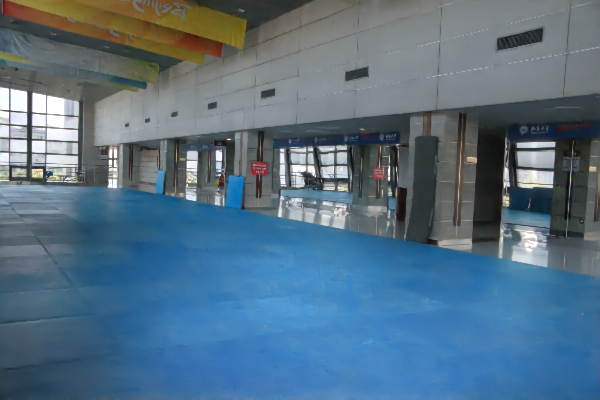}}&
\hspace{-0.42cm}
{\includegraphics[width=2.68cm]{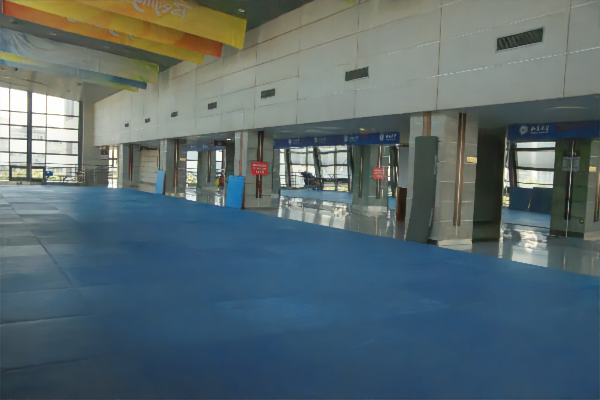}}&
\hspace{-0.42cm}
{\includegraphics[width=2.68cm]{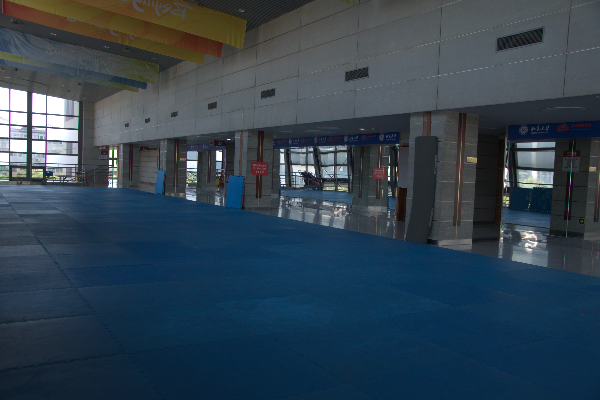}}\\
\hspace{-0.3cm} (a) Original&\hspace{-0.42cm}(b) LIME&\hspace{-0.42cm}(c) RetinexNet&\hspace{-0.42cm}(d) DRBN&\hspace{-0.42cm}(e) Ours&\hspace{-0.42cm}(f) Reference
\end{tabular}
\caption{The visual results of state-of-the-art methods and ours on real low-light images from LOL-Real test dataset (top-to-bottom).}
\label{compare}
\end{figure}
%Quantitative Evaluation
\subsection{Comparison with Existing Methods}
We conduct extensive experiments using real instances from the LOL-Real dataset \cite{yang2021sparse}. Results are compared between R2-MWCNN and the other different state-of-the-art methods, including both classic and deep learning-based. Representative results are visually shown in Figure \ref{compare}. In general, most previous methods implemented in the pixel domain have limited denoising and structure recovery capacities. By checking the details, BPDHE, LIME perform well in streching contrast. Nevertheless, blind noise is heavily amplified, resulting in severe visual degradation. The great performance of DRBN, CRM and SICE in the objective evaluation shows their ability to restore global illumination and visibility. However, structural information and image details are not promising. Comparatively, R2-MWCNN can detach both the additive and non-additive noise from the image structural information in the frequency domain effectively. In this way, our method achieves good detail fidelity and perceptual visual quality, as well as satisfactory global illumination. 
%E-GAN denotes EnlightenGAN.
\begin{table}[htbp]
\footnotesize
	\centering
	\caption{Quantitative results in {\em LOL-Real} dataset \cite{yang2021sparse}. (Red: The best, Blue: The second)}
	\label{quantitative}
	%\resizebox{\textwidth}{!}{
	\begin{tabular}{ c|cccccc }
		\hline
		Metric &BIMEF \cite{2017bio} &  BPDHE \cite{ibrahim2007brightness} & CRM \cite{ying2017new} &  RetinexNet \cite{LOL} &  EFF \cite{ying2017} & LDR \\
			\hline
			\hline
			PSNR  &  17.85   & 13.84 & 19.65 &  16.10  &  17.85 & 17.17 \\
			 SSIM &  0.6353  & 0.3990 & 0.6344 & 0.5689  & 0.6353 & 0.5697 \\
			FID &  1.5623    &  2.7219 & 0.6480 &  1.8457 & 1.5623 & 2.7266\\
			\hline
              Metric  & JED \cite{ren2018joint} & LIME \cite{LIME}  & MF \cite{fu2016fusion} &  RRM \cite{li2018structure} & E-GAN \cite{enlightengan} & DONG\\
			\hline
			\hline
 			PSNR  & 17.23 & 15.24  &  18.73   & 17.35 & 18.64 & 17.26\\
  		      SSIM & 0.7256 & 0.4181 &  0.5050  & 0.7251 & 0.6786 & 0.4755\\
			FID  & 1.8785 & 11.908  &  0.9072 &  1.7040 & 0.7902 & 1.1610\\
			\hline
		Metric &  ASMR & MSR \cite{jobson1997multiscale} &  DeepUPE \cite{wang2019}  & SICE \cite{SICE} & DRBN \cite{yang2020fidelity} & R2-MWCNN \\
			\hline
			\hline
			PSNR  & 15.37 &11.67 & 13.17 &  19.40 &  \textcolor{blue}{20.11} &\textcolor{red}{23.23}\\	
			SSIM & 0.4310 & 0.4064 & 0.4509 &  0.6842 &  \textcolor{blue}{0.8495} & \textcolor{red}{0.8648} \\
			FID   & 2.3692 & 7.8184 & 1.1641 & 0.6233 &\textcolor{blue}{0.5039} & \textcolor{red}{0.2716} \\
			\hline
	\end{tabular}
\label{objective}
\end{table}

Table \ref{objective} contains a quantitative comparison of various methods. We adopt three objective image quality representatives as performance criteria: Structure Similarity (SSIM), Peak Signal-to-Noise Ratio (PSNR) and Frechet Inception Distance (FID) \cite{heusel2017gans}. PSNR is utilized to measure the detail fidelity between the enhanced image and the normal light image. SSIM is more concerned with structural information and global illumination recovery. Consider that the perceptual visual quality might not be well captured by the PSNR and SSIM. We utilize FID score to measure the similarity between the enhanced image and the groundtruth image from the high-level context. The lower FID value indicates a greater quantity of similarity. Our method (Red) outperforms all the other methods in PSNR, SSIM and FID by a large margin. More concretely, R2-MWCNN considerably surpasses the second best method (Blue) in all three criteria, which means our method is equipped with better denoising, structural information restoration and can achieve very good visual fidelity.

% i.e. pixel  by turning one or more of them off at one time
\subsection{Ablation Study}
In this section, we analyse the effect of each technical design to support our motivation.  Several ablation studies have been conducted to demonstrate that our method is promising to achieve satisfactory enhanced results. Table \ref{as} shows the objective results (PSNR/SSIM) of the MWCNN and the R2-MWCNN with different loss function combinations.
\begin{table}[htbp]
\footnotesize
	\centering
	\caption{Ablation study of proposed methods on R2-MWCNN and MWCNN \cite{liu2018}\\}
	\label{ablation}
	%\resizebox{\textwidth}{!}{
	\begin{tabular}{ c||c|c|c|c||c|c }
		\hline
		Metric & $L_{pixel}$ & $L_{global}$ & $L_{edge}$ & $L_{channel}$ &  PSNR(dB) &  SSIM \\
			\hline
			\hline
			MWCNN [4]  & \Checkmark  &  \Checkmark & \XSolid & \XSolid&  19.8476 &  0.8390  \\
			 & \Checkmark  &  \Checkmark &  \Checkmark & \XSolid& 20.9474 & 0.8494\\
			 & \Checkmark  &  \Checkmark &   \XSolid   &\Checkmark & 21.4889  & 0.8478\\
			%& \Checkmark  &  \Checkmark & \Checkmark  &  \Checkmark & \XSolid& 22.0322 & 0.8572\\
			& \Checkmark  &  \Checkmark &   \Checkmark & \Checkmark  &\textcolor{red}{ 22.5147}  & \textcolor{red}{ 0.8529}\\
			\hline
			R2-MWCNN & \Checkmark  &  \Checkmark &  \XSolid & \XSolid & 20.9446 & 0.8517 \\
			 & \Checkmark  &  \Checkmark &  \Checkmark & \XSolid &21.8730 & 0.8544\\
			& \Checkmark  &  \Checkmark  & \XSolid   &\Checkmark & 22.3621 & 0.8512 \\		
 			 % & \Checkmark  &  \Checkmark & \Checkmark  & \Checkmark & \XSolid & 22.8303& 0.8620\\
			& \Checkmark  &  \Checkmark &  \Checkmark & \Checkmark  & \textcolor{red}{23.2260} & \textcolor{red}{0.8648}\\

			\hline
	\end{tabular}
\label{as}
\end{table}

In each experiment, we turn off one or more of the loss functions to study their role in the enhancement. The same combination comparison between R2-MWCNN and MWCNN reveals that R2-MWCNN performs better in denoising and detail fidelity. In each block, the first row with previous loss functions serves as the baseline for comparison.
The second and third rows serve as separate verifications of the proposed methods. The last row results show that the network with our loss functions achieves the best low-light enhancement performance.

% 这一优势要在实验的ablation study中体现。
%The goal of this loss is to suppress color disbalance, which typically produced by over or under enhancement of one of the RGB channels after low-light images have been enhanced. The channel-wise loss can learn the distribution of color from high-light photos to guide low-light images augmentation.

\section{Conclusion}
%仿照abstract进行修改
In this paper, we design a frequency-based framework for low-light image enhancement. DWT is introduced to make the network learn to remove noise from the structural information. The proposed MSC module can adaptively select low-/high-level semantic information for satisfactory detail reconstruction. We have also presented a channel-wise loss for color correction in low-light enhancement. Finally, we have conducted extensive experiments to verify the effectiveness of our method against state-of-the-art methods.

% These properties make our method superior to existing methods, and  evaluations demonstrate that our method outperforms the state-of-the-arts by a large margin.
%-------------------------------------------------------------------------

\bibliography{egbib}

\begin{thebibliography}{10}

\bibitem{wade1998fast}
Alex~Robert Wade and Frederick~W Fitzke.
\newblock A fast, robust pattern recognition system for low light level image
  registration and its application to retinal imaging.
\newblock {\em Optics Express}, 3(5):190--197, 1998.

\bibitem{rashed2019fusemodnet}
Hazem Rashed, Mohamed Ramzy, Victor Vaquero, Ahmad El~Sallab, Ganesh Sistu, and
  Senthil Yogamani.
\newblock Fusemodnet: Real-time camera and lidar based moving object detection
  for robust low-light autonomous driving.
\newblock In {\em Proceedings of the IEEE/CVF International Conference on
  Computer Vision Workshops}, pages 0--0, 2019.

\bibitem{HE}
Stephen~M. Pizer, E.~Philip Amburn, John~D. Austin, Robert Cromartie, Ari
  Geselowitz, Trey Greer, Bart {ter Haar Romeny}, John~B. Zimmerman, and Karel
  Zuiderveld.
\newblock Adaptive histogram equalization and its variations.
\newblock {\em Computer Vision, Graphics, and Image Processing},
  39(3):355--368, 1987.

\bibitem{Land}
Edwin~H. Land.
\newblock The retinex theory of color vision.
\newblock {\em Scientific American}, 237(6):108--129, 1977.

\bibitem{cheng2004simple}
Heng-Da Cheng and XJ~Shi.
\newblock A simple and effective histogram equalization approach to image
  enhancement.
\newblock {\em Digital signal processing}, 14(2):158--170, 2004.

\bibitem{abdullah2007dynamic}
Mohammad Abdullah-Al-Wadud, Md~Hasanul Kabir, M~Ali~Akber Dewan, and Oksam
  Chae.
\newblock A dynamic histogram equalization for image contrast enhancement.
\newblock {\em IEEE Transactions on Consumer Electronics}, 53(2):593--600,
  2007.

\bibitem{wang2013naturalness}
Shuhang Wang, Jin Zheng, Hai-Miao Hu, and Bo~Li.
\newblock Naturalness preserved enhancement algorithm for non-uniform
  illumination images.
\newblock {\em IEEE Transactions on Image Processing}, 22(9):3538--3548, 2013.

\bibitem{LOL}
Chen Wei, Wenjing Wang, Wenhan Yang, and Jiaying Liu.
\newblock Deep retinex decomposition for low-light enhancement.
\newblock {\em arXiv preprint arXiv:1808.04560}, 2018.

\bibitem{yang2021sparse}
Wenhan Yang, Wenjing Wang, Haofeng Huang, Shiqi Wang, and Jiaying Liu.
\newblock Sparse gradient regularized deep retinex network for robust low-light
  image enhancement.
\newblock {\em IEEE Transactions on Image Processing}, 30:2072--2086, 2021.

\bibitem{LIME}
Xiaojie Guo, Yu~Li, and Haibin Ling.
\newblock Lime: Low-light image enhancement via illumination map estimation.
\newblock {\em IEEE Transactions on Image Processing}, 26(2):982--993, 2017.

\bibitem{yang2020fidelity}
Wenhan Yang, Shiqi Wang, Yuming Fang, Yue Wang, and Jiaying Liu.
\newblock From fidelity to perceptual quality: A semi-supervised approach for
  low-light image enhancement.
\newblock In {\em Proceedings of the IEEE/CVF Conference on Computer Vision and
  Pattern Recognition}, pages 3063--3072, 2020.

\bibitem{cai2018learning}
Jianrui Cai, Shuhang Gu, and Lei Zhang.
\newblock Learning a deep single image contrast enhancer from multi-exposure
  images.
\newblock {\em IEEE Transactions on Image Processing}, 27(4):2049--2062, 2018.

\bibitem{chen2018learning}
Chen Chen, Qifeng Chen, Jia Xu, and Vladlen Koltun.
\newblock Learning to see in the dark.
\newblock In {\em Proceedings of the IEEE Conference on Computer Vision and
  Pattern Recognition}, pages 3291--3300, 2018.

\bibitem{guo2020zero}
Chunle Guo, Chongyi Li, Jichang Guo, Chen~Change Loy, Junhui Hou, Sam Kwong,
  and Runmin Cong.
\newblock Zero-reference deep curve estimation for low-light image enhancement.
\newblock In {\em Proceedings of the IEEE/CVF Conference on Computer Vision and
  Pattern Recognition}, pages 1780--1789, 2020.

\bibitem{liu2021pd}
Yijun Liu, Zhengning Wang, Yi~Zeng, Hao Zeng, and Deming Zhao.
\newblock Pd-gan: Perceptual-details gan for extremely noisy low light image
  enhancement.
\newblock In {\em ICASSP 2021-2021 IEEE International Conference on Acoustics,
  Speech and Signal Processing (ICASSP)}, pages 1840--1844. IEEE, 2021.

\bibitem{agaian2001transform}
Sos~S Agaian, Karen Panetta, and Artyom~M Grigoryan.
\newblock Transform-based image enhancement algorithms with performance
  measure.
\newblock {\em IEEE Transactions on image processing}, 10(3):367--382, 2001.

\bibitem{bijalwan2012wavelet}
Akhilesh Bijalwan, Aditya Goyal, and Nidhi Sethi.
\newblock Wavelet transform based image denoise using threshold approaches.
\newblock {\em International Journal of Engineering and Advanced Technology
  (IJEAT) ISSN}, 1:2249, 2012.

\bibitem{liu2018}
Pengju Liu, Hongzhi Zhang, Kai Zhang, Liang Lin, and Wangmeng Zuo.
\newblock Multi-level wavelet-cnn for image restoration.
\newblock In {\em Proceedings of the IEEE conference on computer vision and
  pattern recognition workshops}, pages 773--782, 2018.

\bibitem{clahe}
Ali~M Reza.
\newblock Realization of the contrast limited adaptive histogram equalization
  (clahe) for real-time image enhancement.
\newblock {\em Journal of VLSI signal processing systems for signal, image and
  video technology}, 38(1):35--44, 2004.

\bibitem{ibrahim2007brightness}
Haidi Ibrahim and Nicholas Sia~Pik Kong.
\newblock Brightness preserving dynamic histogram equalization for image
  contrast enhancement.
\newblock {\em IEEE Transactions on Consumer Electronics}, 53(4):1752--1758,
  2007.

\bibitem{lee2013contrast}
Chulwoo Lee, Chul Lee, and Chang-Su Kim.
\newblock Contrast enhancement based on layered difference representation of 2d
  histograms.
\newblock {\em IEEE transactions on image processing}, 22(12):5372--5384, 2013.

\bibitem{nakai2013color}
Keita Nakai, Yoshikatsu Hoshi, and Akira Taguchi.
\newblock Color image contrast enhacement method based on differential
  intensity/saturation gray-levels histograms.
\newblock In {\em 2013 International Symposium on Intelligent Signal Processing
  and Communication Systems}, pages 445--449. IEEE, 2013.

\bibitem{jobson1997properties}
Daniel~J Jobson, Zia-ur Rahman, and Glenn~A Woodell.
\newblock Properties and performance of a center/surround retinex.
\newblock {\em IEEE transactions on image processing}, 6(3):451--462, 1997.

\bibitem{rahman1996multi}
Zia-ur Rahman, Daniel~J Jobson, and Glenn~A Woodell.
\newblock Multi-scale retinex for color image enhancement.
\newblock In {\em Proceedings of 3rd IEEE International Conference on Image
  Processing}, volume~3, pages 1003--1006. IEEE, 1996.

\bibitem{lee2013adaptive}
Chang-Hsing Lee, Jau-Ling Shih, Cheng-Chang Lien, and Chin-Chuan Han.
\newblock Adaptive multiscale retinex for image contrast enhancement.
\newblock In {\em 2013 International Conference on Signal-Image Technology \&
  Internet-Based Systems}, pages 43--50. IEEE, 2013.

\bibitem{jobson1997multiscale}
Daniel~J Jobson, Zia-ur Rahman, and Glenn~A Woodell.
\newblock A multiscale retinex for bridging the gap between color images and
  the human observation of scenes.
\newblock {\em IEEE Transactions on Image processing}, 6(7):965--976, 1997.

\bibitem{li2018structure}
Mading Li, Jiaying Liu, Wenhan Yang, Xiaoyan Sun, and Zongming Guo.
\newblock Structure-revealing low-light image enhancement via robust retinex
  model.
\newblock {\em IEEE Transactions on Image Processing}, 27(6):2828--2841, 2018.

\bibitem{fu2016weighted}
Xueyang Fu, Delu Zeng, Yue Huang, Xiao-Ping Zhang, and Xinghao Ding.
\newblock A weighted variational model for simultaneous reflectance and
  illumination estimation.
\newblock In {\em Proceedings of the IEEE Conference on Computer Vision and
  Pattern Recognition}, pages 2782--2790, 2016.

\bibitem{llnet}
Kin~Gwn Lore, Adedotun Akintayo, and Soumik Sarkar.
\newblock Llnet: A deep autoencoder approach to natural low-light image
  enhancement.
\newblock {\em Pattern Recognition}, 61:650--662, 2017.

\bibitem{shen2017msr}
Liang Shen, Zihan Yue, Fan Feng, Quan Chen, Shihao Liu, and Jie Ma.
\newblock Msr-net: Low-light image enhancement using deep convolutional
  network.
\newblock {\em arXiv preprint arXiv:1711.02488}, 2017.

\bibitem{enlightengan}
Yifan Jiang, Xinyu Gong, Ding Liu, Yu~Cheng, Chen Fang, Xiaohui Shen, Jianchao
  Yang, Pan Zhou, and Zhangyang Wang.
\newblock Enlightengan: Deep light enhancement without paired supervision.
\newblock {\em IEEE Transactions on Image Processing}, 30:2340--2349, 2021.

\bibitem{wang2019}
Ruixing Wang, Qing Zhang, Chi-Wing Fu, Xiaoyong Shen, Wei-Shi Zheng, and Jiaya
  Jia.
\newblock Underexposed photo enhancement using deep illumination estimation.
\newblock In {\em Proceedings of the IEEE/CVF Conference on Computer Vision and
  Pattern Recognition}, pages 6849--6857, 2019.

\bibitem{SICE}
Jianrui Cai, Shuhang Gu, and Lei Zhang.
\newblock Learning a deep single image contrast enhancer from multi-exposure
  images.
\newblock {\em IEEE Transactions on Image Processing}, 27(4):2049--2062, 2018.

\bibitem{MBLLEN}
Feifan Lv, Feng Lu, Jianhua Wu, and Chongsoon Lim.
\newblock Mbllen: Low-light image/video enhancement using cnns.
\newblock In {\em BMVC}, page 220, 2018.

\bibitem{pan1999two}
Quan Pan, Lei Zhang, Guanzhong Dai, and Hongai Zhang.
\newblock Two denoising methods by wavelet transform.
\newblock {\em IEEE transactions on signal processing}, 47(12):3401--3406,
  1999.

\bibitem{ronneberger2015u}
Olaf Ronneberger, Philipp Fischer, and Thomas Brox.
\newblock U-net: Convolutional networks for biomedical image segmentation.
\newblock In {\em International Conference on Medical image computing and
  computer-assisted intervention}, pages 234--241. Springer, 2015.

\bibitem{xu2020learning}
Ke~Xu, Xin Yang, Baocai Yin, and Rynson~WH Lau.
\newblock Learning to restore low-light images via
  decomposition-and-enhancement.
\newblock In {\em Proceedings of the IEEE/CVF Conference on Computer Vision and
  Pattern Recognition}, pages 2281--2290, 2020.

\bibitem{alom2018recurrent}
Md~Zahangir Alom, Mahmudul Hasan, Chris Yakopcic, Tarek~M Taha, and Vijayan~K
  Asari.
\newblock Recurrent residual convolutional neural network based on u-net
  (r2u-net) for medical image segmentation.
\newblock {\em arXiv preprint arXiv:1802.06955}, 2018.

\bibitem{lu2012learning}
Yao Lu, Wei Zhang, Cheng Jin, and Xiangyang Xue.
\newblock Learning attention map from images.
\newblock In {\em 2012 IEEE Conference on Computer Vision and Pattern
  Recognition}, pages 1067--1074. IEEE, 2012.

\bibitem{venables2013modern}
William~N Venables and Brian~D Ripley.
\newblock {\em Modern applied statistics with S-PLUS}.
\newblock Springer Science \& Business Media, 2013.

\bibitem{fukui2019attention}
Hiroshi Fukui, Tsubasa Hirakawa, Takayoshi Yamashita, and Hironobu Fujiyoshi.
\newblock Attention branch network: Learning of attention mechanism for visual
  explanation.
\newblock In {\em Proceedings of the IEEE/CVF Conference on Computer Vision and
  Pattern Recognition}, pages 10705--10714, 2019.

\bibitem{sara2019image}
Umme Sara, Morium Akter, and Mohammad~Shorif Uddin.
\newblock Image quality assessment through fsim, ssim, mse and psnr—a
  comparative study.
\newblock {\em Journal of Computer and Communications}, 7(3):8--18, 2019.

\bibitem{johnson2016perceptual}
Justin Johnson, Alexandre Alahi, and Li~Fei-Fei.
\newblock Perceptual losses for real-time style transfer and super-resolution.
\newblock In {\em European conference on computer vision}, pages 694--711.
  Springer, 2016.

\bibitem{wang2004image}
Zhou Wang, Alan~C Bovik, Hamid~R Sheikh, and Eero~P Simoncelli.
\newblock Image quality assessment: from error visibility to structural
  similarity.
\newblock {\em IEEE transactions on image processing}, 13(4):600--612, 2004.

\bibitem{Simonyan2015VeryDC}
K.~Simonyan and Andrew Zisserman.
\newblock Very deep convolutional networks for large-scale image recognition.
\newblock {\em CoRR}, abs/1409.1556, 2015.

\bibitem{abadi2016tensorflow}
Mart{\'\i}n Abadi, Ashish Agarwal, Paul Barham, Eugene Brevdo, Zhifeng Chen,
  Craig Citro, Greg~S Corrado, Andy Davis, Jeffrey Dean, Matthieu Devin, et~al.
\newblock Tensorflow: Large-scale machine learning on heterogeneous distributed
  systems.
\newblock {\em arXiv preprint arXiv:1603.04467}, 2016.

\bibitem{2017bio}
Zhenqiang Ying, Ge~Li, and Wen Gao.
\newblock A bio-inspired multi-exposure fusion framework for low-light image
  enhancement.
\newblock {\em arXiv preprint arXiv:1711.00591}, 2017.

\bibitem{ying2017new}
Zhenqiang Ying, Ge~Li, Yurui Ren, Ronggang Wang, and Wenmin Wang.
\newblock A new low-light image enhancement algorithm using camera response
  model.
\newblock In {\em Proceedings of the IEEE International Conference on Computer
  Vision Workshops}, pages 3015--3022, 2017.

\bibitem{ying2017}
Zhenqiang Ying, Ge~Li, Yurui Ren, Ronggang Wang, and Wenmin Wang.
\newblock A new image contrast enhancement algorithm using exposure fusion
  framework.
\newblock In {\em International Conference on Computer Analysis of Images and
  Patterns}, pages 36--46. Springer, 2017.

\bibitem{ren2018joint}
Xutong Ren, Mading Li, Wen-Huang Cheng, and Jiaying Liu.
\newblock Joint enhancement and denoising method via sequential decomposition.
\newblock In {\em 2018 IEEE International Symposium on Circuits and Systems
  (ISCAS)}, pages 1--5. IEEE, 2018.

\bibitem{fu2016fusion}
Xueyang Fu, Delu Zeng, Yue Huang, Yinghao Liao, Xinghao Ding, and John Paisley.
\newblock A fusion-based enhancing method for weakly illuminated images.
\newblock {\em Signal Processing}, 129:82--96, 2016.

\bibitem{heusel2017gans}
Martin Heusel, Hubert Ramsauer, Thomas Unterthiner, Bernhard Nessler, and Sepp
  Hochreiter.
\newblock Gans trained by a two time-scale update rule converge to a local nash
  equilibrium.
\newblock {\em arXiv preprint arXiv:1706.08500}, 2017.

\end{thebibliography}

\end{document}